\documentclass{article}

\usepackage{arxiv}

\usepackage[utf8]{inputenc} 
\usepackage[T1]{fontenc}    
\usepackage{hyperref}       
\usepackage{url}            
\usepackage{booktabs}       
\usepackage{amsfonts}       
\usepackage{nicefrac}       
\usepackage{microtype}      

\usepackage{lipsum}         
\usepackage{graphicx}
\usepackage{natbib}
\usepackage{doi}
\usepackage{makecell}
\usepackage{amssymb,cuted}
\usepackage{mathtools} 
\usepackage{amsmath} 
\usepackage{cleveref}       
\graphicspath{{./images/}} 
\crefdefaultlabelformat{#2\bfseries\upshape(#1)#3}
\creflabelformat{eq.}{#2\bfseries\upshape(#1)#3}

\def\p{{\boldsymbol p}}

\title{In tree structure should sentence be generated}

\author{
	Yaguang Li \\
    Beijing, China \\
	\texttt{arkliyg@gmail.com} \\
    \And
	Xin Chen \\
    London, U.K. \\
	\texttt{xin.chen.uk@gmail.com} \\
}

\renewcommand{\headeright}{}
\renewcommand{\undertitle}{}
\renewcommand{\shorttitle}{In Tree Structure should Sentence be Generated}

\hypersetup{
pdftitle={SenTree}
}

\begin{document}
\maketitle

\begin{abstract}
    Generative models reliant on sequential autoregression have been at the forefront of language generation for an extensive period, particularly following the introduction of widely acclaimed transformers.
    Despite its excellent performance, there are always some issues that we face today.
    For example, problems such as hallucinations and getting trapped in a logic loop may occur.
    To enhance the performance of existing systems, this paper introduces a new method for generating sequences in natural language, which involves generating the targeted sentence in a tree-traversing order.
    The paper includes an illustration of the theoretical basis and validity of the approach, as well as a comparison of its fundamentals with the diffusion model in graphic generation.
    Finally, a module called SenTree is introduced for generating an approximating binary tree.
    It is already available at \url{https://github.com/arklyg/sentree}.
    Additionally, a joint training framework based on this approach is proposed, incorporating the intrinsics of generative adversarial networks.
\end{abstract}

\section{Introduction} \label{sec:intro}
    As a basic language model in which the famous attention mechanism is leveraged, transformers \cite{vaswani2017attention} have achieved state-of-the-art performance in various natural language processing (NLP) tasks, including natural language generation.
    Aside from the marvelous idea of the widely used attention mechanism in the architecture, the way it generating is sometimes questioned, especially the potential low performance of sequential autoregressive generation \cite{cundy2024sequencematch}.

    \Cref{sec:background} introduces the notion that while human thought processes are recursive, our language follows a tree-like structure.
    Traditionally, human speech unfolds in a sequential manner due to the constraints imposed by audio media delivery.
    In contrast, contemporary information processing systems afford the flexibility to convey information in diverse structures beyond one-dimensional representations.
    Given the varying semantic contributions of individual words and the long-term decay property introduced by position embedding, \Cref{sec:background} establishes the theoretical certainty of generating sentences in the order of binary tree traversal.

    The process of generating sentences in a binary tree traversal order resembles the methodology employed in denoising diffusion probabilistic models \cite{ho2020denoising}.
    This comparison is expounded upon in \Cref{sec:intrinsic}, highlighting the approach as a generalization of the diffusion model from graphic to natural language generation.
    Building upon the theory that words with greater significance should be generated earlier, an approximating hierarchical generation method is introduced in \Cref{sec:methods}, yielding a superior BLEU score in translation tasks compared to the original transformer, described in \Cref{sec:results}.
    The method is lightweight and easy to integrate into existing systems based on the widely used transformer.

    In \Cref{sec:conclusion}, we analyze the potential of the proposed approach for generating tree structures, suggesting that there could be a considerable performance enhancement by refining the tree structure generation model based on BERT \cite{devlin2018bert} when the loss of the generative language model is not decreasing.
    The entire process which resembles generative adversarial networks (GANs) \cite{Goodfellow2014GenerativeAN}, seeks a suitable tree generation structure from both perspectives.
    However, while GANs are adversarial, the proposed framework is collaborative.

\section{Background} \label{sec:background}
\subsection{Words with Weight}

    The semantic content of a sentence is not equally distributed among all the words \cite{filippova-strube-2008-dependency}, only a few content words are necessary to convey the main meaning of a sentence in some cases \cite{kauf2024lexical}.
    For instance, the sentence "semantic content not equally distributed, few words convey" summarizes the previous long sentence.
    Typically, only a small number of words in a sentence contribute to its main meaning, while many words such as "of", "but", or "a" have minimal semantic impact.
    Additional negligible words are addressed in \cite{groff1976semantics}.
    The subsequent sections of the paper will discuss the significance of words in a sentence as their "weight.”

    Let $F(S, n)$ be the semantic content generating function, in which, $S$ is a sentence and the $n^{th}$ element is ignored. Let $\mathbb\{w_i\}_{i=1}^{N}$ be a sentence constructed of $N$ tokens with $w_i$ being the $i^{th}$ element, there would be:
\begin{equation}
    \begin{aligned}
        Semantics_n &=F(\{w_i\}_{i=1}^{N}, n),\\
    \end{aligned}
    \label{fn:semantic}
\end{equation}
    And Let $Distance_{\{{w_i}_{i=1}^{N}\}}(Semantics)$ be the distance from $Semantics$ to the original Semantics of $\mathbb\{w_i\}_{i=1}^{N}$. To any sentence, there will be a sequence $n_1$, $n_2$, ... ,$n_N$, which satisfied: 
\begin{equation}
    \begin{aligned}
        Distance_{\{{w_i}_{i=1}^{N}\}}(Semantics_{n_1}) > Distance_{\{{w_i}_{i=1}^{N}\}}(Semantics_{n_2}) >... > Distance_{\{{w_i}_{i=1}^{N}\}}(Semantics_{n_N}),\\
    \end{aligned}
    \label{fn:distance}
\end{equation}
    And there will be a sequence of word weights corresponded:
\begin{equation}
    \begin{aligned}
        Weight_{\{{w_i}_{i=1}^{N}\}}(w_{n_1}) > Weight_{\{{w_i}_{i=1}^{N}\}}(w_{n_2}) > ... > Weight_{\{{w_i}_{i=1}^{N}\}}(w_{n_N}),\\
    \end{aligned}
    \label{fn:weight}
\end{equation}
    We can see the weight of words within a sentence varies.
    It is important to note that the weight of the word is context-dependent \cite{javorsky2023assessing}.

\subsection{Long-term Decay of Probability}
    The current methods of generating sentences prioritize the order of natural language patterns, but they do not adequately consider important words.
    Heavy-weighted words may end up being placed far from their intended position in the generated sentence.

    The probability the word $w_t$ could be generated in a sentence is illustrated as:
\begin{equation}
    \begin{aligned}
        P(w_t) =P(w_t|\{{w_i}_{i=1}^{t-1}\}),\\
    \end{aligned}
    \label{fn:probability}
\end{equation}

    If the word is expected to be generated at position $t$, thus:

\def\approxprop{%
  \def\p{%
    \setbox0=\vbox{\hbox{$\propto$}}%
    \ht0=0.6ex \box0 }%
  \def\s{%
    \vbox{\hbox{$\sim$}}%
  }%
  \mathrel{\raisebox{0.7ex}{%
      \mbox{$\underset{\s}{\p}$}%
    }}%
}

\begin{equation}
    \begin{aligned}
        Attention(\{{w_i}_{i=1}^{t-1}\}, w_t) \approxprop \sum_{i=1}^{t-1} pos\_embedding(w_i)\cdot pos\_embedding(w_t)
    \end{aligned}
    \label{fn:attention}
\end{equation}

\begin{figure}
    \centering
    \includegraphics[width=0.7\textwidth]{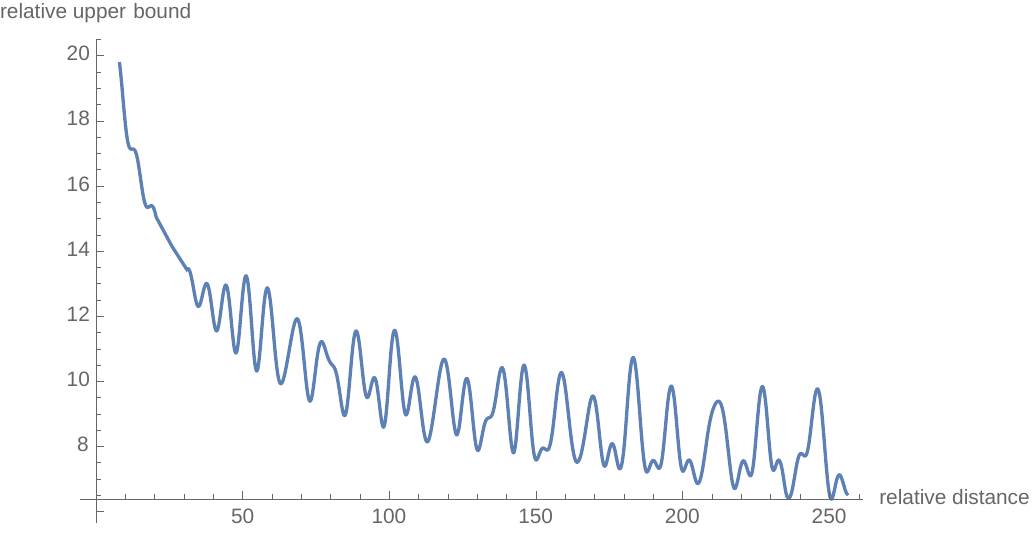}
    \caption{Long-term decay of position embedding (RoPE).}
    \label{fig:decay}
\end{figure}

    Most position embedding algorithms deployed in transformers have a long-term decay property, which means that the inner product will decay when the relative position increases, as shown in \Cref{fig:decay}, \cite{Su_2024}.
    This property coincides with the intuition that a pair of words with a long relative distance should have less connection, which is justifiable and necessary in a language model based on the attention mechanism.

    This property of position embedding algorithms leads us to:
\begin{equation}
    \begin{aligned}
        Attention(\{{w_i}_{i=1}^{t+1}\}, w) \lessapprox Attention(\{{w_i}_{i=1}^{t}\}, w),\\
    \end{aligned}
    \label{fn:attention-decay}
\end{equation}
    That is:
\begin{equation}
    \begin{aligned}
    P(w|\{{w_i}_{i=1}^{t+1}\}) \lessapprox P(w|\{{w_i}_{i=1}^{t}\}),\\
    \end{aligned}
    \label{fn:probability-decay}
\end{equation}
    Which is, after the expected position, the longer an expected word takes to be generated in a sentence, the more difficult it is.
    Moreover, if the word carries significant weight, the sentence may fail to convey the intended meaning and instead create a loop in logic or a semantic trap due to the absence of the expected word.

\subsection{Defects Introduced by Long-term Decay of Probability}
    In modern practices, the defects that appear in LLMs are:
    
\begin{itemize}
    \item If earlier words are not well generated and are lightweight in the designated semantic content, the latter ones with heavy weight are hard to generate, and the semantics will drift and be misled into an awkward loop or trap.
    \item To jump out of the loop or trap, a new prompt is introduced, which brings the outer words to let the state of the generating process back to the right track and lures the model to generate words with heavy weight as soon as expected.
    \item However, the existing generation method is confined to human language sequential expressions, so words with heavy weight would not be generated at a considerable chance as expected.
\end{itemize}

    Defects should not be attributed to the long-term decay property of position embedding algorithms because this property is essential.
    What we need to improve is the sequential generation process.

\section{Intrinsics}\label{sec:intrinsic}
\subsection{Heavier Words First}
    To accurately convey the intended meaning of a given sentence, we should place the word with heavier weight first and the word with lesser weight later, which is:
\begin{equation}
    \begin{aligned}
        min(Distance_{\{{w_i}_{i=1}^{N}\}}(Semantics_0)) \propto - log[P(x_1)P(x_2|x_1)\prod \limits_{n=3}^NP(x_n|x_1, x_2, ..., x_{n-1})],\\
    where\ \ x_n\in \{{w_i}_{i=1}^{N}\}\ \ and\ \ Weight_{\{{w_i}_{i=1}^{N}\}}(x_{n}) > Weight_{\{{w_i}_{i=1}^{N}\}}(x_{n+1}),\\
    \end{aligned}
    \label{fn:probability-decay}
\end{equation}
    So, if we generate the sentence in this manner, we can maximize the probability of approaching the intended semantics.
\subsection{Binary Tree}
    To preserve the original positional information, one effective method is to use a binary tree as the sentence representation.
    This is because a binary tree possesses the following properties:
\begin{itemize}
    \item It implies order information.
    \item It is hierarchically structured and can represent the relative weight of words.
    \item It is equivalent to a definite sequence, which is compatible with the output of transformers.
\end{itemize}

    Compared to using a tree structure approach, using a raw sentence approach pays more effort for sentence completion rather than building semantics.
    This approach is sensitive to variations in the latest generated word and may deviate from the original intended semantics.
    On the other hand, the tree structure approach is more stable.
    In a tree structure, words close to the "root" carry more meaning and set the tone of the sentence, while words farther from the "root" can be substituted in various ways or not generated at all.

    Once the main word is determined, the foundation of the sentence is established and an appropriate expression is on its way.
    The words near the "root" generated in the early stage help establish the main meaning, which will change less in the next stage of generation.
    This is because the subsequent word can be substituted with many others without changing the meaning significantly.
    The "tree" refers to the syntax tree in the study of semantics, but in the transformer model, there is no concept of syntax, only the attention mechanism.

\begin{figure}
    \centering
    \includegraphics[width=0.7\textwidth]{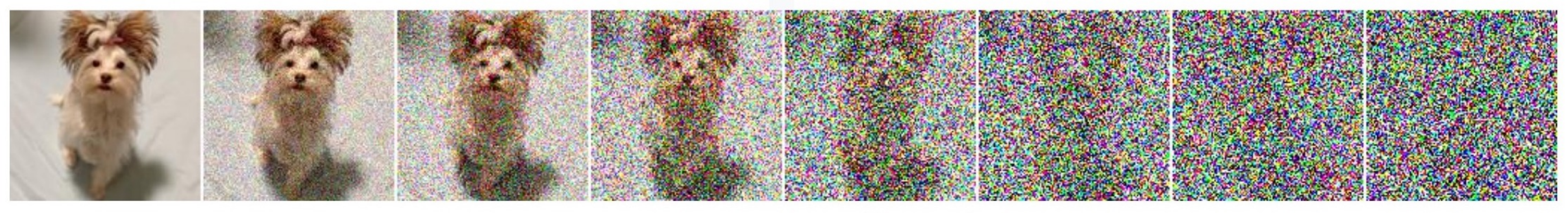}
    \caption{Adding noise process of diffusion model.}
    \label{fig:diffusion}
\end{figure}

\subsection{Correlations with Diffusion Model}
    The diffusion \cite{ho2020denoising} models also serve as a great source of inspiration.
    It begins with total noise and gradually becomes a clear image.
    In the reverse process, it starts as an aggregating noise process.
    As the process continues, the more vague parts merge with the noise first, while the less vague parts merge later.
    Important characters are hard to dilute when adding noise and can remain distinguishable in later steps.
    This is illustrated in \Cref{fig:diffusion}  \cite{song2021scorebased}.
    Eventually, we can see a slightly blurred contour or a few characters in the noise, emerging from the total noise image.
    In the reverse process of adding noise, providing the appropriate blurred contour or few characters in the first step is important because all future completion steps are based on these initial elements.

    Also, the process of generating a sentence can be thought of as a de-noising process in one-dimensional space initially filled with complete noise until the first word emerges.
    Therefore, it's crucial to place more emphasis on important words in earlier steps when de-noising.
    This means that the sequence of generated words should sort from closer to farther from the intended semantic meaning, which is the weight of words mentioned in this paper.

    In the process of generating a tree, each step only provides a word and its relative position.
    The exact placement of the word is uncertain until the very last word is generated.
    This means that the distribution of every word in any step of the generating process forms a range where each word may take several consecutive possible positions at any step.
    An interesting finding is that these ranges of possible positions are always overlap until the very last word generated, creating a spectrum across the entire one-dimensional space, showing a similar form of generating content with the diffusion models.

\begin{figure}
    \centering
    \includegraphics[width=0.7\textwidth]{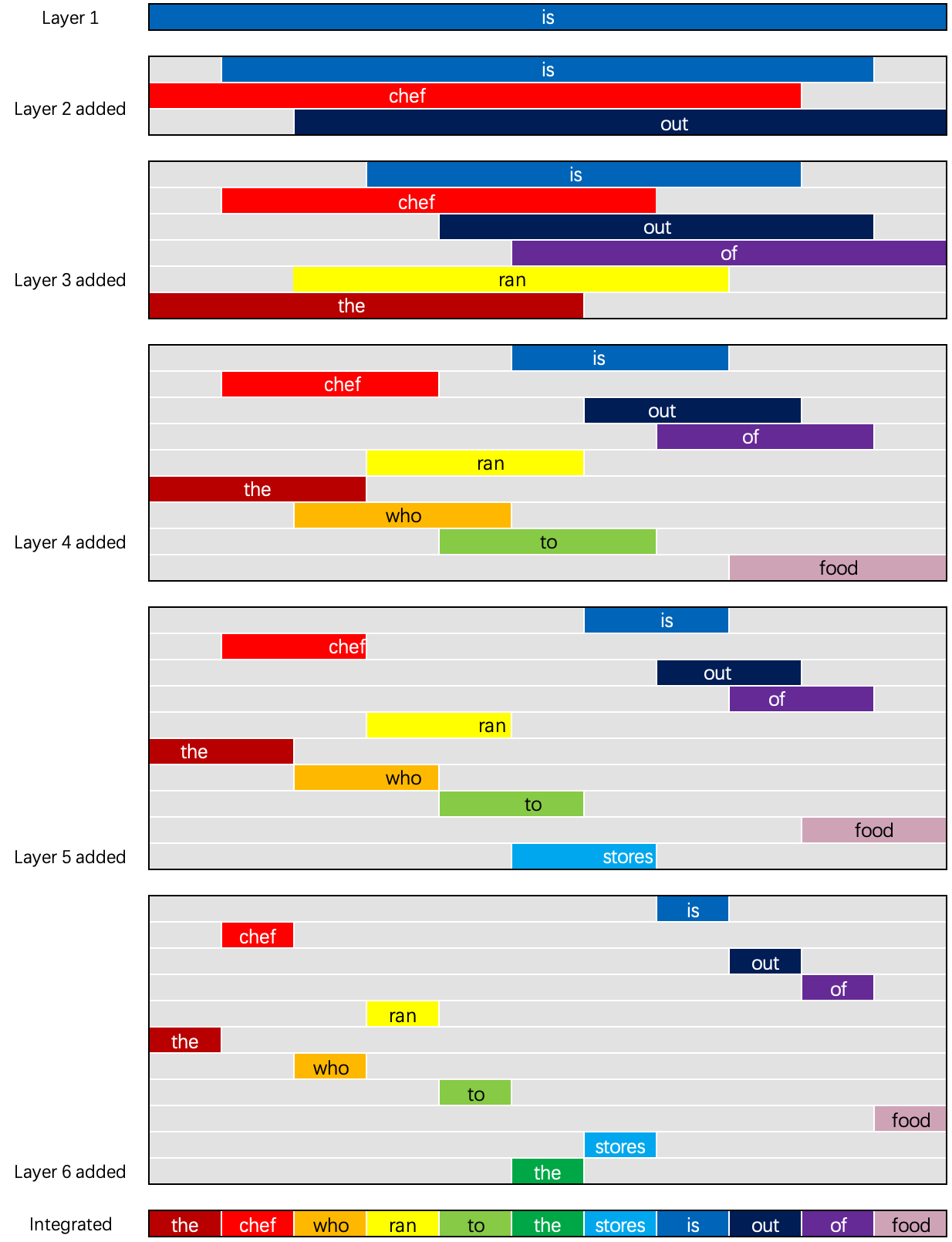}
    \caption{Tree generating process representing in spectra.}
    \label{fig:tree-diffusion}
\end{figure}

\begin{figure}
    \centering
    \includegraphics[width=0.7\textwidth]{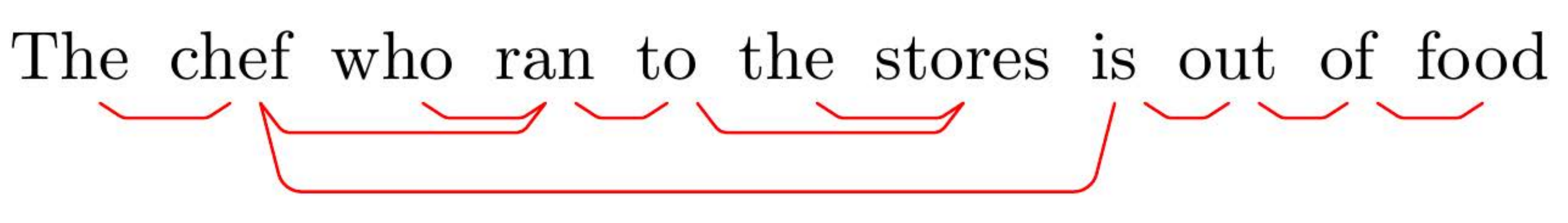}
    \caption{Structural tree of sentence.}
    \label{fig:structural-sentence}
\end{figure}

    As depicted in \Cref{fig:tree-diffusion}, the graph illustrates the process of the binary tree structuring a sentence being generated by adding words of each layer, with the actual sentence tree shown in \Cref{fig:structural-sentence}, \cite{hewitt-manning-2019-structural}.
    Spectra represents the range of possible positions where a word could be placed, and this range diminishes as new layers of words added.
    After the last layer of words is added, the range diminishes to a definitive position where the word is finally located.
    The generating process goes from an "opaque" tree to a "clear" one, similar to the diffusion model's de-noising process, while which is carried out in two-dimensional space.

\section{Methods}\label{sec:methods}
\subsection{Learnable Models rather than Fixed Algorithms}
    The use of algorithms to convert a sentence into a tree structure compliant with language syntax has been a topic of interest.
    However, it's important to consider the limitations of algorithms in this context.
    Firstly, the term "tree" typically refers to the syntax tree in semantic studies, but in the transformer model, there is no concept of syntax, only the attention mechanism.
    Therefore, there is no need to generate a tree with algorithms adhering to syntax rules.
    Secondly, the sentence-tree converter is expected to improve as more data is fed into it and becomes more suitable for generative models, whereas fixed algorithms may not adapt well.

    While the tree structure may not perfectly align with the human understanding of a syntax tree, it is sufficient to represent how a machine parses a sentence in the context of attention.
    The attention mechanism in natural language processing performs a similar function, allowing the machine to learn the information structure of a sentence.
    This is different from the human-created syntax tree, although both involve parsing a sentence into a tree structure.
    Therefore, the tree structure learned through the attention mechanism is more suitable for representing information, especially in the context of information structuring.

\subsection{Structural-Probe}
    Due to advancements in auto-encoding language models, there are now several models capable of converting sentences into binary trees.
    In \cite{hewitt-manning-2019-structural}, a structural probe was introduced to generate the syntax tree, which represents the tree structure that the generative language model produces.
    The previous research on sentence tree structure suggests that the structural probe, based on the BERT model \cite{devlin2018bert}, can be used to achieve this structure.
    In the structural-probe study, a linear transformation $B$ was trained to establish the tree distance between all pairs of words in sentences from a parsed corpus.
    The researchers found that the sentence dependency tree is indeed embedded in language models like BERT \cite{devlin2018bert} or Elmo \cite{elmo}.
    They refer to the transformation as a structural probe, which provides a specific assertion about the structure of the vector space.
\begin{figure}
    \centering
    \includegraphics[width=0.7\textwidth]{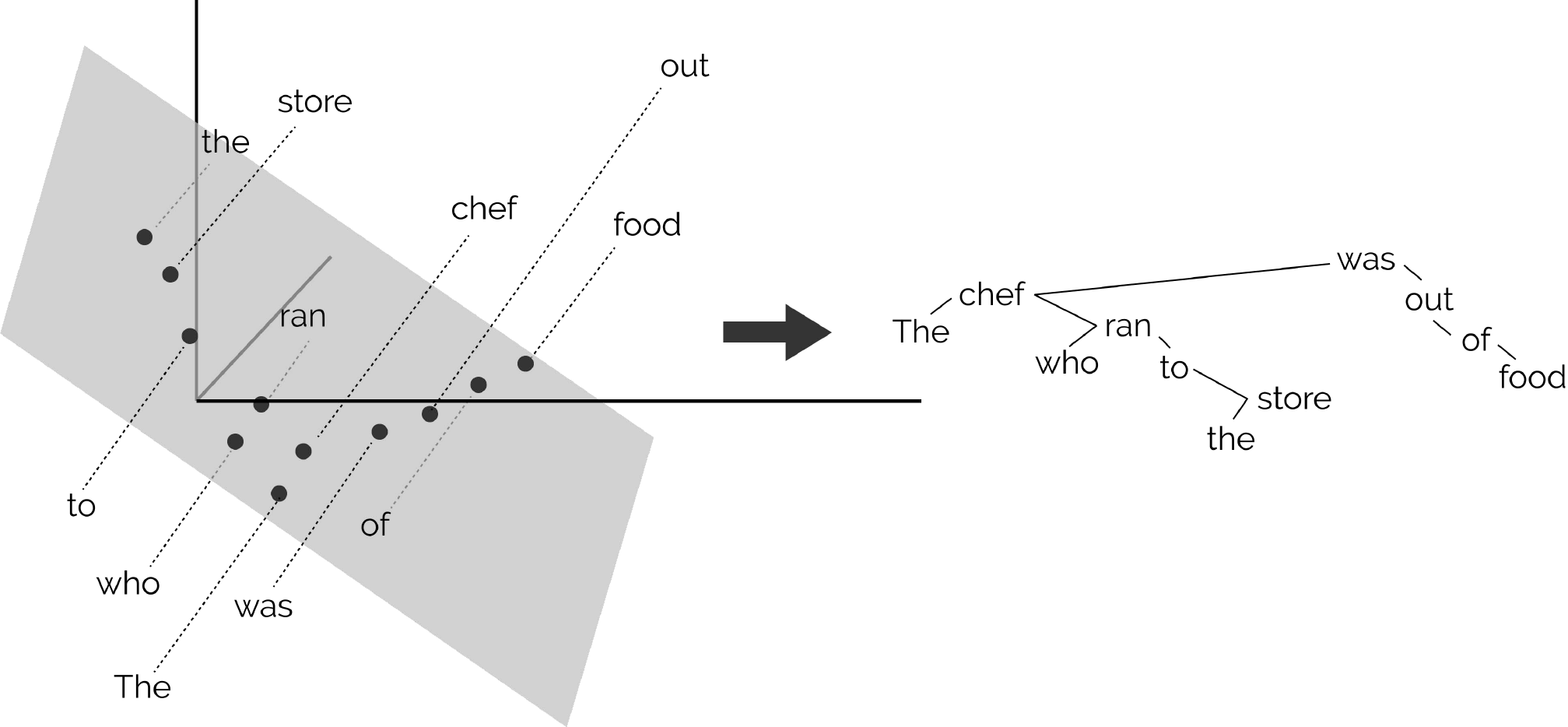}
    \caption{Transformation by structural-probe based on BERT}
    \label{fig:structural-probe}
\end{figure}
    The transformation is depicted in \Cref{fig:structural-probe} \cite{hewitt-manning-2019-structural}.

\subsection{Process}
    To begin with, the structural-probe is used to determine the depth of each word.
    Using these depths, the word at the top is treated as the root, with the left child being the shallowest word on the left, and similarly on the right.
    This process creates a binary tree with sentree. 

    Next, sentree converts the binary tree into a sequence that can be used as input and output for the decoder.
    Root-first traversal is used, which is a proximating way of traversing the binary tree layer by layer.
    Special tokens such as '\emph{<ITN>}' and '\emph{<VAC>}' are used to denote internal nodes and vacancies.

    Finally, the sequential sentence implied by the binary tree representation of the transformer output can be converted back into a sequential sentence by sentree. 
\begin{figure}
    \centering
    \includegraphics[width=0.7\textwidth]{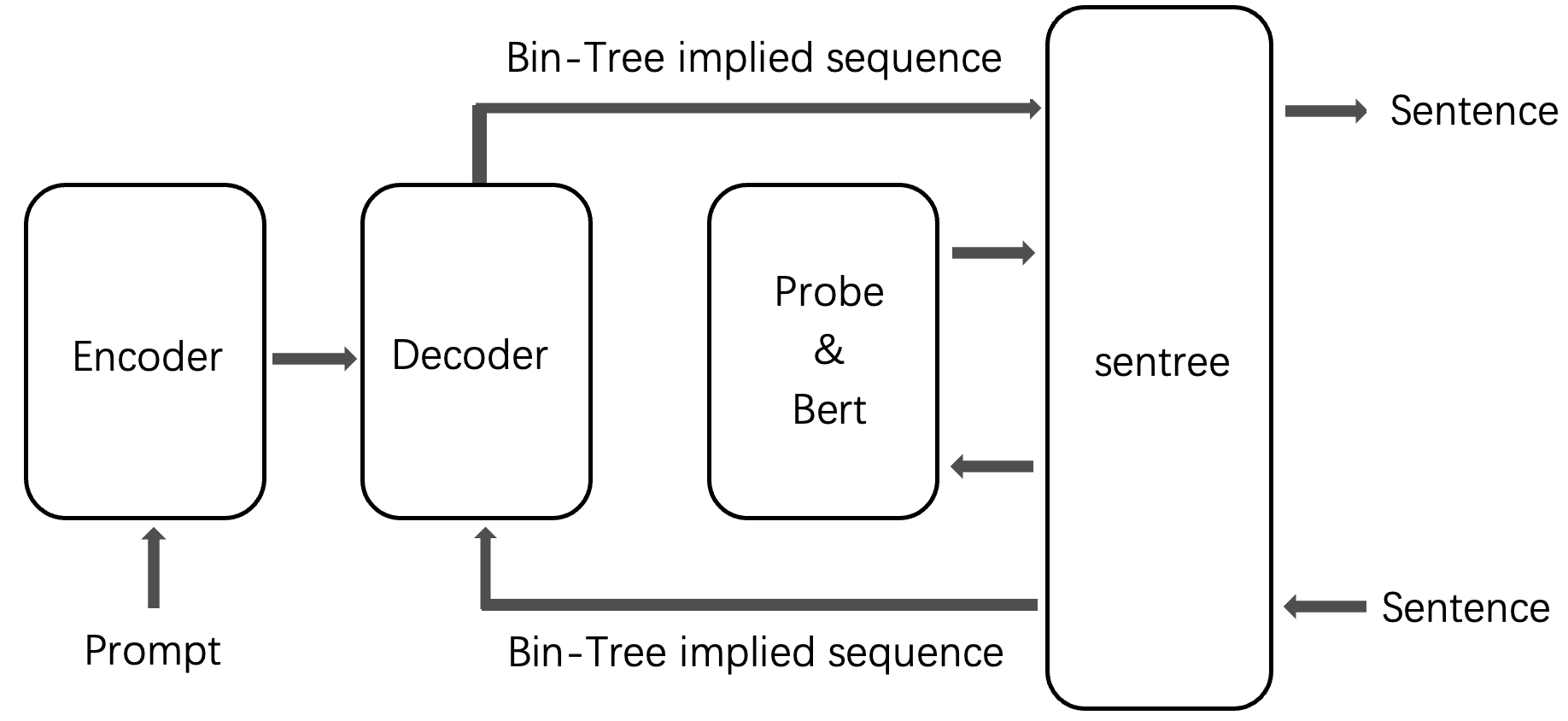}
    \caption{Transformer-Sentree framework}
    \label{fig:sentree-transformer}
\end{figure}
    The entire process is illustrated in \Cref{fig:sentree-transformer}.
    
\section{Experiments and Evaluation}\label{sec:results}

    The BERT \cite{devlin2018bert} based structural-probe \cite{hewitt-manning-2019-structural} is used to generate the binary tree from the targeted sentence.
    A converter - SenTree has been developed to convert a sentence into a binary tree sequential representation, and it also supports the reverse conversion, backed by the structural-probe.
    The converter is available at \url{https://github.com/arklyg/sentree}.
    All experiments were conducted using the original encoder-decoder transformer framework \cite{vaswani2017attention} where both the encoder and decoder consist of a stack of $N = 6$ identical layers, producing outputs of dimension $d_{model} = 512$.
    Both models were trained under identical conditions and resource consumption.
    Results are reported in \Cref{tb:mt}, and as we can see, the proposed approach gives better BLEU scores compared to its baseline alternative \cite{vaswani2017attention} on the WMT 2014 English-to-German translation task \cite{mtdataset}.

\begin{table}
	\label{tb:mt}
	\centering
	\begin{tabular}{ll}
		\toprule
		Approach     & BLEU \\
		\midrule
		Transformer-Base & 26.9\\
		Transformer-Sentree & \textbf{27.1} \\ 
		\bottomrule
	\end{tabular}
	\caption{Results comparison between approches of Transformer-base and Transformer-sentree.}
\end{table}

\section{Conclusions and Future Work}\label{sec:conclusion}
    In this work, a new method for generating sentences that utilizes a learnable binary tree structure converter based on the structural-probe  \cite{hewitt-manning-2019-structural} is proposed, which in turn is based on BERT \cite{devlin2018bert}.
    This novel approach, integrating Transformers and SenTree, has shown improved performance compared to the original sentence-to-sentence generating approach.

    The process of formulating and expressing thoughts is an interesting aspect of human cognition.
    We conceive an idea first and then struggle to articulate it, as we have navigated the complexities of language in advance.
    In large language models (LLMs), these two tasks overlap, leading to certain flaws known as language-thought conflation fallacies \cite{mahowald2024dissociating}.

    The remainder of this paper tries to separate the two tasks by having the encoder prompt understanding and reasoning while the decoder and the sentence-tree converter, backed by an auto-encoding language model, expressing the message.
    After the autoregressive model was trained to a certain limit, following the performance, we can fine-tune the auto-encoding model, on which the converter was built, to match the binary tree implied sequence output of our autoregressive model.
    The model-backed converter and decoder collaborate to find a proper expression in binary tree structure to match the information from the encoder output to the decoder.
    The tree structure generated from a sentence will not be static anymore since the model from which the tree structure is generated could be trained according to the performance of the autoregressive model.
    This approach resembles the well-known generative adversarial networks (GANs) \cite{Goodfellow2014GenerativeAN}.
    Nevertheless, while GANs play a min-max game, the approach involves a win-win one.

    While training the converter:
\begin{equation}
    \begin{aligned}
        loss(G(x))=loss(T(prompt, G(x)),\\
    \end{aligned}
    \label{fn:loss}
\end{equation}
    Where $T(encoder\_input, decoder\_input)$ is as Transformers which encoder input are sentences while decoder input are binary tree implied sequences.
    $G(sentence)$ is the sentence-tree converter, and $x$ is targeted sentences from training data corresponding to prompt sentences $prompt$.

\begin{figure}
    \centering
    \includegraphics[width=0.9\textwidth]{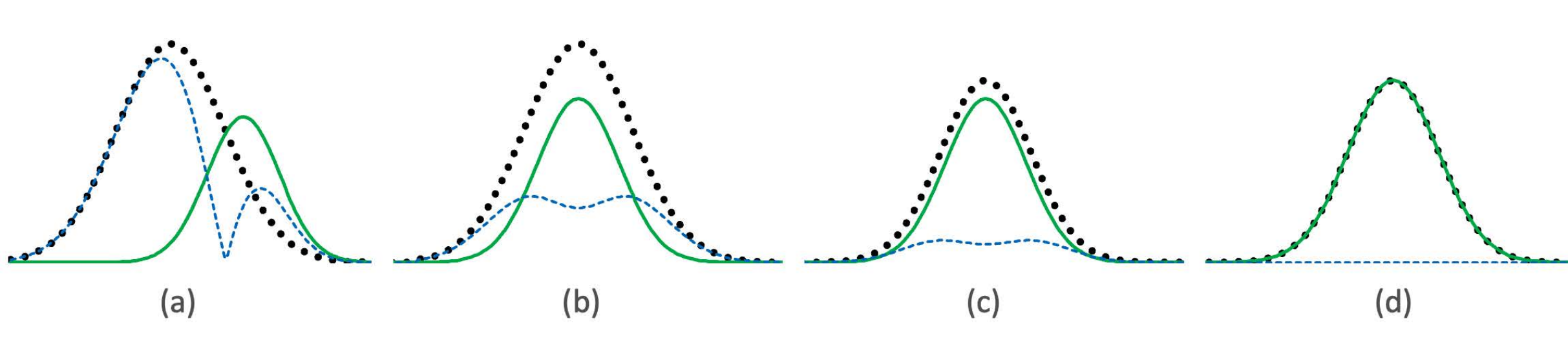}
    \caption{Joint training of transformer and tree-generator}
    \label{fig:joint-training}
\end{figure}

    The joint training process is illustrated in \Cref{fig:joint-training}.
    In which, transformer and tree-generator are trained jointly.
\begin{itemize}
    \item (a) The distribution of $T$ (green, solid line) is initially far away from the distribution of $G$ (black, dotted line), and the loss of theirs is significantly high (blue, dashed line).
    \item (b) We train $T$ first, and the distribution of it tends to converge with that of $G$. The loss decreased to an extent.
    \item (c) Then we train $G$, and the distribution of it tends to converge with that of $T$. The loss decreased to an extent.
    \item (d) We repeat the alternate training process, at last the distributions nearly converge and the loss may become significantly low.
\end{itemize}
    
    Theoretically, this Transformers-SenTree framework can be generalized to other media formats, such as image, voice, video, etc.
    It is believed that there will be easier to finetune, naturally integrated, and more powerful sentence-tree converter proposed, which will lift the performance of autoregressive generative language models to a new level.

\bibliographystyle{unsrtnat}
\bibliography{references}  

\end{document}